# RL-STPA: Adapting System-Theoretic Hazard Analysis for Safety-Critical Reinforcement Learning

**Steven A. Senczyszyn[1], Timothy C. Havens[1], Nathaniel Rice[1],**
**Jason E. Summers[2], Benjamin D. Werner[3], Benjamin J. Schumeg[3],**

[1]Michigan Technological University

[2]ARiA

[3]DEVCOM

sasenczy@mtu.edu, thavens@mtu.edu

**Abstract**

As reinforcement learning (RL) deployments expand into safety-critical domains, existing evaluation methods fail to systematically identify hazards arising from the black-box nature of neural network enabled policies and distributional shift between training and deployment. This paper introduces Reinforcement Learning System-Theoretic Process Analysis (RL-STPA), a framework that adapts conventional STPA's systematic hazard analysis to address RL's unique challenges through three key contributions: hierarchical subtask decomposition using both temporal phase analysis and domain expertise to capture emergent behaviors, coverage-guided perturbation testing that explores the sensitivity of state-action spaces, and iterative checkpoints that feed identified hazards back into training through reward shaping and curriculum design. We demonstrate RL-STPA in the safety-critical test case of autonomous drone navigation and landing, revealing potential loss scenarios that can be missed by standard RL evaluations. The proposed framework provides practitioners with a toolkit for systematic hazard analysis, quantitative metrics for safety coverage assessment, and actionable guidelines for establishing operational safety bounds. While RL-STPA cannot provide formal guarantees for arbitrary neural policies, it offers a practical methodology for systematically evaluating and improving RL safety and robustness in safety-critical applications where exhaustive verification methods remain intractable.

## 1. Introduction

The adoption of reinforcement learning (RL) in safety-critical applications, from autonomous vehicles (Kiran et al. 2021) to healthcare (Yu et al. 2021), has created an urgent need for systematic safety evaluation methodologies. While RL has demonstrated success in complex control tasks, the gap between simulated performance and real-world safety remains a fundamental challenge. Traditional safety analysis methods, designed for interpretable control systems with explicit loss scenarios, struggle to address the opaque decision-making processes of neural network enabled policies.

### 1.1 The Safety Verification Gap

Traditional control systems decompose complex tasks into modular components, such as perception, path planning, and motor control. Each module can be independently verified, with well-defined objectives and loss scenarios. Safety analysis techniques like Fault Tree Analysis (FTA) and Failure Mode and Effects Analysis (FMEA) leverage this modularity to systematically identify hazards.

In contrast, RL agents learn direct mappings from observations to actions through trial-and-error optimization. This end-to-end learning produces policies that may achieve superior performance but lack the granularity and interpretability necessary for traditional safety analysis. The RL agent's decision-making process, encoded in neural network parameters, is difficult to decompose into analyzable components. This "black-box" nature creates several safety challenges:

- **Emergent behaviors**: Complex behaviors emerge from the interaction of learned features that have no explicit representation.
- **Distributional shift**: Policies trained in simulation may exhibit unexpected behaviors when deployed in environments with different statistical properties.
- **Hidden loss scenarios**: Critical failures may occur in state-space regions unexplored during training.
- **Lack of safety constraints**: Standard RL optimization focuses on maximizing reward without explicit consideration of worst-case scenarios.

### 1.2 Why STPA for RL?

System-Theoretic Process Analysis (STPA), developed as part of the System-Theoretic Accident Model and Processes (STAMP) framework, offers a promising foundation for RL safety analysis. Unlike component-focused methods, STPA views safety as a control problem arising from system-level interactions (Leveson 2012). This systems-thinking aligns naturally with RL's holistic approach to control. One important feature of STPA is that it is a worst-case scenario analysis method, which is crucial to understand when deploying RL systems for safety-critical applications, such as military operations and autonomous vehicles. It should also be noted that hazards are conditions that could lead to mishaps/accidents.

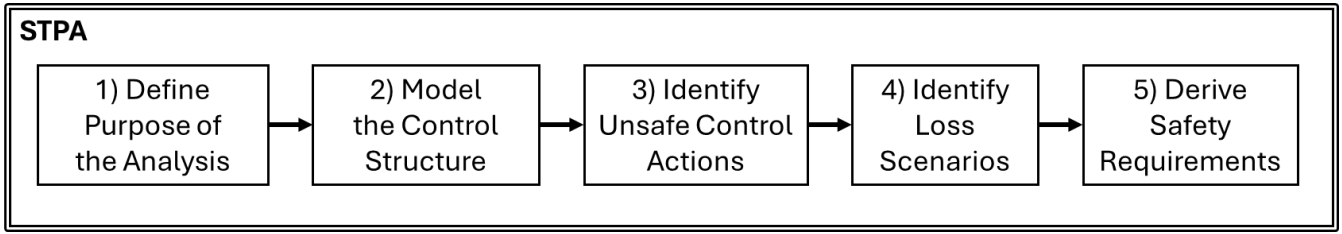

Figure 1: Overview of the STPA hazard analysis workflow.

However, applying STPA to RL systems presents unique challenges. Traditional STPA assumes the ability to identify and analyze individual control actions and their contexts. In RL, the agent's control logic is distributed across a neural network, making it difficult to isolate the decision-making process behind specific control actions. A naive application of STPA to an RL agent would attempt to analyze the entire state-action space simultaneously, which quickly becomes computationally intractable for continuous domains.

Not to be confused with STPA-RL (Chang, Kwon, and Kwon 2024), which implements RL into traditional STPA to improve the identification of loss scenarios, RL-STPA reformulates the STPA process for learning enabled systems that cannot be decomposed into control elements.

### 1.3 Contributions

This paper introduces RL-STPA, a framework that adapts STPA's systematic hazard analysis for the unique challenges of RL. Our key contributions include:

1. **Hierarchical Subtask Decomposition**: We propose a methodology for decomposing monolithic RL policies into analyzable subtasks based on temporal mission phases. This decomposition enables tractable hazard analysis while preserving the emergent behaviors that make RL effective.
2. **Coverage-Guided Perturbation Testing**: We develop a systematic approach to perturbation testing that explores critical regions of the state-action space. By performing grid search over perturbation parameters with importance sampling, we efficiently identify loss scenarios while managing computational costs.
3. **Iterative Checkpoints**: We establish a framework for feeding identified hazards back into the RL development process through reward function modifications and curriculum learning adjustments. This creates an iterative improvement cycle between hazard analysis and RL agent training.

We demonstrate RL-STPA through a detailed case study of autonomous drone navigation and landing, showing how the framework has the potential to reveal safety-critical insights that standard RL evaluation might miss. While our current implementation requires manual interpretation of results, we outline paths toward automation and formal verification.

## 2. Background

### 2.1 System-Theoretic Process Analysis

STPA, developed as a methodology of the STAMP framework by Nancy Leveson (2012) and further refined in Leveson and Thomas (2018), represents a paradigm shift in safety analysis. Rather than viewing accidents as chains of component failures, STPA treats them as results of inadequate control enforcement within complex systems. This perspective proves particularly relevant for RL systems, where emergent behaviors arise from learned control policies. One important feature of STPA is that it is a worst-case scenario analysis method

The STPA process, depicted in Figure 1, consists of five key steps:

1. **Define the Purpose of the Analysis**: Identify losses to prevent, define system-level hazard, and establish system boundaries.
2. **Model the Control Structure**: Create a hierarchical control diagram showing control actions and feedback.
3. **Identify Unsafe Control Actions (UCAs)**: Systematically examine how control actions could lead to hazards.
4. **Identify Loss Scenarios**: Determine why UCAs might occur.
5. **Derive Safety Requirements:** Define constraints to prevent UCAs.

For traditional systems, these steps map clearly to engineered components. For RL systems, we must reinterpret the steps to handle trained neural control policies, which we further describe in Section 3.

### 2.2 Reinforcement Learning Fundamentals

In RL, an agent interacts with an environment, which provides an observation and reward $r$ at each step, to learn an optimal policy $\pi^*$ that maps states $s$ to actions $a$ by maximizing the expected cumulative reward for all time steps $t$ that are discounted by a factor of $\gamma$. This is formalized below in Equation 1 and Figure 2.

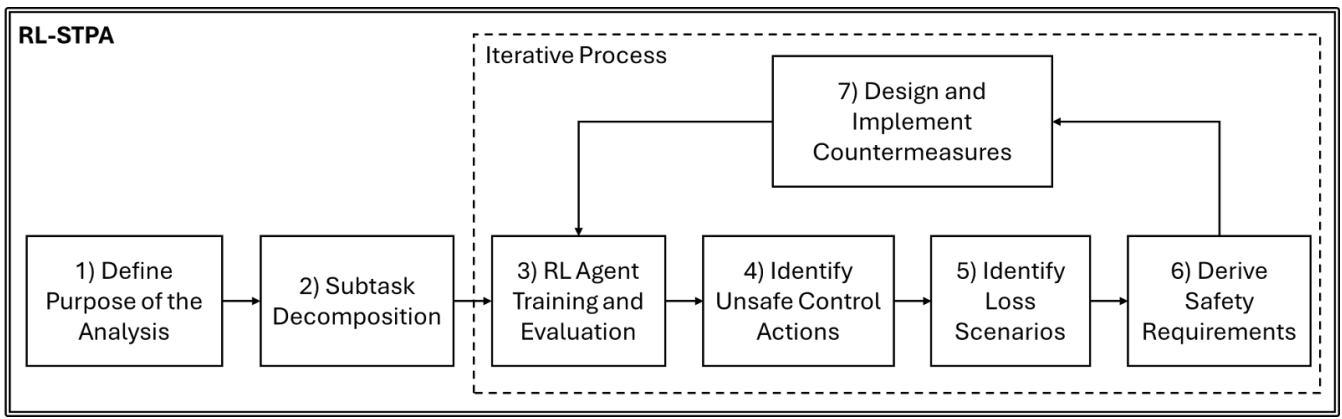

Figure 3: RL-STPA process overview.

$$\pi^* = \arg\max E\left[\sum_{t=0}^{\infty} \gamma^t r_t \mid \pi \right] \quad (1)$$

Modern deep RL algorithms like Proximal Policy Optimization (PPO) or Soft Actor-Critic (SAC) parameterize $\pi$ using neural networks (Sutton and Barto 2004).

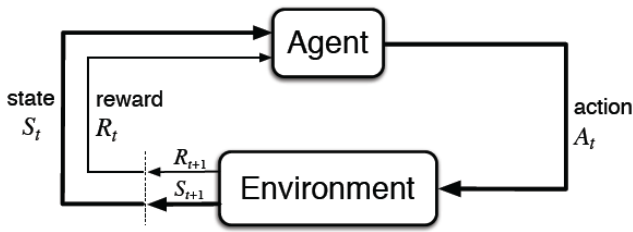

Figure 2: Iterative feedback loop for RL illustrating the interaction between agent and environment

This learning paradigm creates several characteristics relevant to safety analysis:

- **Implicit Feature Learning:** The RL agent automatically discovers relevant state features without manual engineering.
- **Continuous Adaptation:** Policies can be updated based on new experience.
- **Reward Specification Criticality:** The reward function implicitly defines all objectives, including safety.

Importantly, while RL agents are typically trained with a single reward function encompassing all objectives, they can be evaluated using different portions of the reward function post-training. This flexibility enables our subtask decomposition approach.

### 2.3 Related Work in RL Safety

The RL safety community has developed several approaches to address risks associated with safety-critical deployments:

- **Safe Reinforcement Learning:** Methods like Constrained Policy Optimization (Achiam et al. 2017) and PPO-Lagrangian (Ray, Amodei, and Achiam 2019) explicitly incorporate safety constraints during training. These approaches bound the probability of constraint violations but require predefined safety specifications.
- **Formal Verification:** Researchers have adapted neural network verification techniques to provide guarantees about neural policies (Katz et al. 2017) along with formal methods (Fulton and Platzer 2018). However, these methods face severe scalability challenges and typically require simplified policy representations.
- **Shielding and Runtime Monitoring:** Shield synthesis creates safety wrappers that override potentially unsafe actions (Alsheik et al. 2018). While effective for known hazards, shields struggle with emergent behaviors not anticipated during design.
- **Traditional Drone Safety Analysis:** Classical control systems undergo rigorous hazard analysis using methods like Fault Tree Analysis (FTA), Event Tree Analysis (ETA), and Hazard and Operability Study (HAZOPS) (Leveson 2012). These analyses decompose the system into subsystems (navigation, stability control, obstacle avoidance) each with distinct functions and loss scenarios.

RL-STPA differs from these approaches by providing a systematic methodology for discovering unknown hazards through the lens of system-theoretic thinking, without requiring formal specifications or simplified models.

## 3. RL-STPA Methodology

The RL-STPA framework adapts traditional STPA to address the unique challenges of analyzing learned policies. We maintain STPA's systems-theoretic perspective while introducing novel techniques for handling RL's black-box nature.

The process begins with a definition of the intended purpose of the analysis and a trained RL agent. It then iteratively refines our understanding of both its loss scenarios and the RL agent itself through targeted retraining. Figure 3 illustrates the complete RL-STPA workflow, and each step is explained in further detail. While STPA itself may be an iterative process, RL-STPA has an additional iterative process within itself to develop and implement countermeasures that aim to satisfy safety requirements. Experimental methods are utilized to find loss scenarios rather than analyzing the control structure.

### 3.1 Define the Purpose of the Analysis

The initial phase of RL-STPA remains unchanged to that of STPA. In this phase the primary goal is to define the losses that the analysis aims to prevent, the hazards that lead to those losses, and the system-level constraints that need to be satisfied. Additionally, operating requirements such as security, performance, and other system properties must be defined. This includes objectives such as preventing loss of human life, ensuring mission success, and prevention of property damage.

There are three basic criteria for defining system-level hazards according to Leveson and Thomas (2018): hazards are system states or conditions (not component-level causes or environmental states), hazards will lead to a loss in some worst-case environment, and hazards must describe states or conditions to be prevented. Hazards should be states to be prevented and states that we never want the system to be in, not states that the system must normally be in to accomplish its goals. Each identified hazard is then linked back to the losses it is associated with.

System-level constraints are then defined that specifies system conditions or behaviors that need to be satisfied to prevent hazards (and ultimately losses).

### 3.2 Subtask Decomposition

The primary difference between STPA and RL-STPA manifests in how the system control structure is modeled. Unlike STPA where traditional control systems are modeled, RL-STPA requires a different approach due to the end-to-end learning approach of RL which results in a black-box control model. Therefore, a radically different approach must be taken, and to achieve this we present the hierarchical subtask decomposition of the mission structure. This involves temporally analyzing the phases involved in achieving the mission objectives and organizing them into subtasks for individual evaluation.

Traditional RL training optimizes a single, monolithic reward function that captures all desired behaviors. This holistic optimization allows the RL agent to discover complex strategies that balance multiple objectives. However, it obscures how the RL agent handles individual aspects of the task. One of the key insights is that while RL agents must be trained monolithically to capture emergent behaviors, they can be evaluated on isolated subtasks based on the temporal phase of the mission. This phase separation enables focused safety analysis without sacrificing the benefits of end-to-end learning.

Subtask decomposition transforms a whole-system analysis into manageable components. We identify subtasks through:

1. **Temporal Phase Analysis:** Dividing missions into sequential phases.
2. **Behavioral Clustering:** Grouping states where the policy exhibits similar activation patterns.
3. **Domain Expert Input:** Leveraging human understanding of natural task divisions.

For each identified subtask, we create isolated test scenarios that activate the relevant portion of the learned policy. This isolation enables targeted hazard analysis and perturbation testing while preserving the policy's learned representations.

### 3.3 RL Agent Training and Evaluation

The RL agent undergoes initial training on the end-to-end process using a comprehensive reward function that captures both task objectives and safety constraints. The training and evaluation proceed in two phases:

1. **Baseline Training:** The RL agent is trained using domain randomization across key environmental parameters for a fixed number of time steps.
2. **Evaluation Protocol:** Post-training evaluation systematically tests the RL agent across each subtask.

For each evaluation, we track both task-specific metrics (e.g. success rate, efficiency) and safety metrics (e.g. minimum obstacle clearance, maximum acceleration, control smoothness).

### 3.4 Identify Unsafe Control Actions

After the mission objectives have been decomposed into subtasks for evaluation, a comprehensive evaluation phase is conducted to identify the Unsafe Control Actions (UCAs). UCAs are defined as a control action that, in a particular context and worst-case environment, will lead to a hazard. In this section, the UCAs are defined for each control action, however due to the black-box nature of RL individual control actions are not necessarily specified. In the context of RL, a more suitable term might be Unsafe Agent Actions, however for consistency we will continue using UCAs.

With that context in mind, following STPA principles, we identify UCAs by analyzing how the RL agent's actions in each subtask could lead to system-level hazards. For RL systems, UCAs often manifest as:

- **Incorrect action magnitude:** Excessive or insufficient control inputs.
- **Timing violations:** Delayed or premature actions.
- **Missing actions:** Failure to respond to critical situations.
- **Inappropriate actions:** Correct type but wrong context.

For each identified UCA, we trace causal factors through the learned policy. While we cannot fully interpret neural network policy decisions, we can identify statistical correlations between environmental conditions and unsafe behaviors.

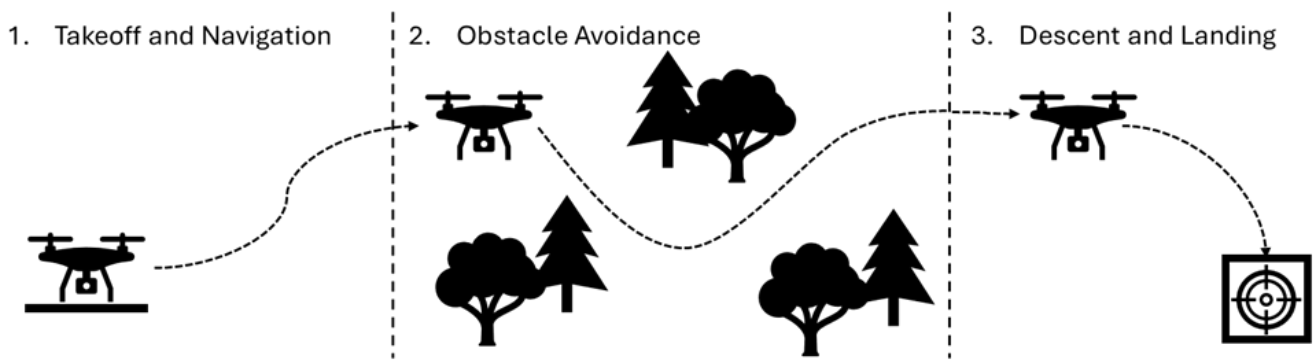

Figure 4: Subtask decomposition of drone navigation and landing mission.

### 3.5 Identify Loss Scenarios

Based on the subtasks and associated UCAs that have been identified, the loss scenarios, or failure modes, are investigated using the results of the evaluation trajectories for the given subtask. This step aims to quantify the likelihood of each loss scenario and the severity of the UCAs. The primary objective in this phase is to understand why the UCAs occur, the likelihood of them occurring, and why control actions would be improperly executed or not executed, leading to hazards.

Real-world deployment introduces perturbations absent from training, such as sensor noise, actuator delays, and environmental disturbances. We systematically explore the policy's robustness through grid search over perturbation parameters. A systematic approach is employed to identify loss scenarios:

**Trajectory Analysis Pipeline:**
1. Evaluate trajectories to identify common failure patterns.
2. Apply interpretability techniques to identify critical decision points
3. Correlate environmental conditions with loss scenarios.

**Perturbation Categories:**
- **Sensor degradation:** Gaussian noise ($\sigma \in [0, 0.5]$), occlusion (0-50% field of view), delay (0-5 frames)
- **Actuator faults:** Response lag (0-5 time steps), Gaussian noise ($\sigma \in [0, 0.5]$)
- **Environmental:** Wind (0-40 mph), visibility (fog density 0-1), dynamic obstacles

### 3.6 Derive Safety Requirements

This step translates identified hazards into actionable improvements. We generate two types of output:
1. **Operational Safety Bounds**: Define the validated operational envelope where the RL agent performs safely. These bounds inform deployment constraints and runtime monitoring systems.
2. **Training Modifications**: Adjustments to the reward function and training curriculum to address identified weaknesses. For example, if landing exhibits high susceptibility to wind perturbations, we might increase wind variation in training scenarios, add penalty terms for excessive touchdown velocity, and/or implement curriculum learning that gradually increases wind intensity.

### 3.7 Design and Implement Countermeasures

Once the loss scenarios have been identified, the proper countermeasures are designed and implemented to improve the RL training process to meet the mission objectives. This creates an iterative process that then leads back to the second phase of subtask decomposition where the implemented countermeasures are then reevaluated so that the UCAs can be redefined along with the loss scenarios. This process is then repeated until the loss scenarios are mitigated to a level that meets the mission objectives. This creates an iterative cycle where safety analysis directly improves the RL agent's robustness. If the evaluation meets all the safety criteria, then this process terminates.

## 4. Application to Autonomous Drone

In this section, RL-STPA is demonstrated through a comprehensive hazard analysis of an autonomous RL agent trained for drone navigation and landing. This application highlights the framework's practical utility along with its potential to improve the training protocol and uncover loss scenarios.

### 4.1 System Description and Mission Profile

First, the mission details are outlined. In this case, a quadrotor unmanned aerial vehicle (UAV) is tasked with autonomous package delivery in a forest environment. The mission profile consists of launching from a designated pad, navigating through open airspace, navigating through the forest and avoiding obstacles, and delivering the payload to target location. The drone is modeled after the Crazyflie 2.0 equipped with LiDAR, GPS, and IMU. The operating requirements provided limit the maximum velocity of the drone to 15 m/s, the minimum separation from obstacles to 0.25 m, and the payload must be delivered within 0.5 m of the target location. Then, based on the given scenario and operating requirements, the relevant losses, hazards, and system-level constraints are identified, which are listed below:

**Identified Losses:**

[L-1]  Death or injury of a person
[L-2]  Property damage
[L-3]  Drone is damaged or destroyed
[L-4]  Loss of mission objective

**Identified Hazards:**

[H-1]  Drone violates minimum separation from obstacles (environmental or other aircraft)
[H-2]  Drone fails to reach target destination
[H-3]  Drone fails to recover from excessive maneuver

**System-Level Constraints:**

[SC-1] Drone must satisfy a minimum separation of 0.25m from obstacles encountered
[SC-2] If the drone violates minimum separation, then the violation must be detected and measures taken to prevent collision

### 4.2 Subtask Decomposition

Once the mission objectives and requirements have been established, temporal phase analysis is applied to successful mission trajectories. The following three primary subtasks were identified, which are also visualized in Figure 4:

1. Takeoff and point-to-point navigation
2. Obstacle avoidance
3. Descent and landing

### 4.3 Agent Training and Evaluation Results

The RL agent then undergoes initial training using a comprehensive reward function that captures mission objectives and basic safety constraints. The RL agent is trained to learn the entire end-to-end process of navigation and landing. In this scenario, the PPO algorithm (Schulman et al. 2017) is used, which is an actor-critic type algorithm. There are two networks involved, the actor chooses an action at each step, and the critic evaluates the reward. Training is then performed for 20,000,000 time steps using a reward function which is derived from the works of Jiang and Song (2022) and Polvara et al. (2018). Before the full training run was initialized, the hyperparameters were tuned via Bayesian optimization using Optuna (Akiba et al. 2019) over 50 trials. The resulting hyperparameters used during training are listed in Table 1. For performance evaluation, we track the primary task-specific metrics of success rate and safety metric of minimum obstacle separation.

The RL agent is then evaluated on the three subtasks that were previously identified. The results presented will focus on the second subtask of obstacle avoidance. The obstacle avoidance subtask is designed so that the trained RL agent starts at a fixed distance from the obstacle with the target location on the opposite side and must reach the same distance on the opposite side of the obstacle. The baseline performance of the obstacle avoidance subtask is then evaluated over 20 episodes with the trajectories recorded for each episode. Once the baseline characterization is complete, the RL agent is then evaluated on the subtask with perturbations applied. The RL agent was then evaluated using a custom penalty function to only characterize the current subtask performance, detailed in Equation 2, where $d_{thresh}$ is the distance threshold for minimum separation and $d_{lidar}$ is the current distance between the drone and the obstacle as measured by the LiDAR sensor.

$$Penalty = -5 * (d_{thresh} - d_{lidar})/d_{thresh} \quad (2)$$

| Hyperparameter | Value |
|---|---|
| Learning Rate | 0.0003 |
| Step Size | 2048 |
| Batch Size | 256 |
| Epochs | 10 |
| Gamma | 0.99 |
| GAE Lambda | 0.95 |
| Clip Range | 0.2 |

Table 1: Optimized hyperparameters used during training.

To illustrate perturbation testing, we conducted perturbation testing over varying levels of wind and obstacle configurations for the obstacle avoidance subtask. Table 2 presents the corresponding RL agent success rate under increasing levels of wind.

| Wind Level | Success Rate |
|---|---|
| No Wind (0.0 m/s) | 100% |
| Light Breeze (2.25 m/s) | 100% |
| Light Wind (4.5 m/s) | 95% |
| Medium Wind (9.0 m/s) | 55% |
| Heavy Wind (18.0 m/s) | 40% |

Table 2: Success rate of obstacle avoidance subtask relative to increasing levels of environmental wind.

### 4.4 Identification of Unsafe Control Actions

Based on the identified subtasks, the UCAs are then defined for each control task associated with that subtask. Because STPA is the worst-case scenario analysis, we assume that if a UCA occurs it will subsequently lead to the associated hazard. Table 3 details the UCAs that were identified for the obstacle avoidance subtask and provide a description of the UCA that occurs based on the control action violation and the associated hazards.

### 4.5 Identification of Loss Scenarios

Once the UCAs were identified for the obstacle avoidance control action, the loss scenarios were then evaluated to understand why the UCAs occurred. The trajectories of the drone performing obstacle avoidance subtask over 20 rollouts for the baseline scenario with a single tree obstacle and no wind, a three tree cluster obstacle and no wind, and a single tree obstacle with medium wind are shown in Figure 5. These trajectories were then used to identify the loss scenarios and UCAs that caused them.

| Control Action | Obstacle Avoidance |
|---|---|
| Not Providing Causes Hazard | UCA-1: RL agent fails to provide Obstacle Avoidance control action when approaching critical environmental obstacles [H-1] |
| Providing Causes Hazard | UCA-2: RL agent provides Obstacle Avoidance control action when no obstacle is present [H-3] |
| Too Early, Too Late, Out of Order | UCA-3: RL agent fails to provide Obstacle Avoidance control action with sufficient time to safely avoid the obstacle [H-1] |
| Stopped Too Soon, Applied Too Long | UCA-4: RL agent fails to apply the Obstacle Avoidance action with sufficient magnitude to avoid obstacles with safe distance [H-1] |

Table 3: Identified UCAs for obstacle avoidance control action associated with drone navigation and landing.

Based on the resulting trajectories, the loss scenarios can be observed for each of the three scenarios. In the baseline scenario with a single tree obstacle and no wind, the RL agent successfully performs obstacle avoidance during all trajectories, however the minimum separation is violated but does not lead to a loss scenario. In the second scenario, with a three tree cluster and no wind, the success rate is reduced to 90% and the minimum separation is violated several times. This loss scenario can be attributed to UCA-4, where the RL agent does not apply obstacle avoidance control action with sufficient magnitude. In the third scenario evaluated, which contains the single tree obstacle under medium wind, the RL agent performs UCAs that lead to loss scenarios, and the success rate is reduced to 55%. This is primarily attributed again to UCA-4, improper magnitude,

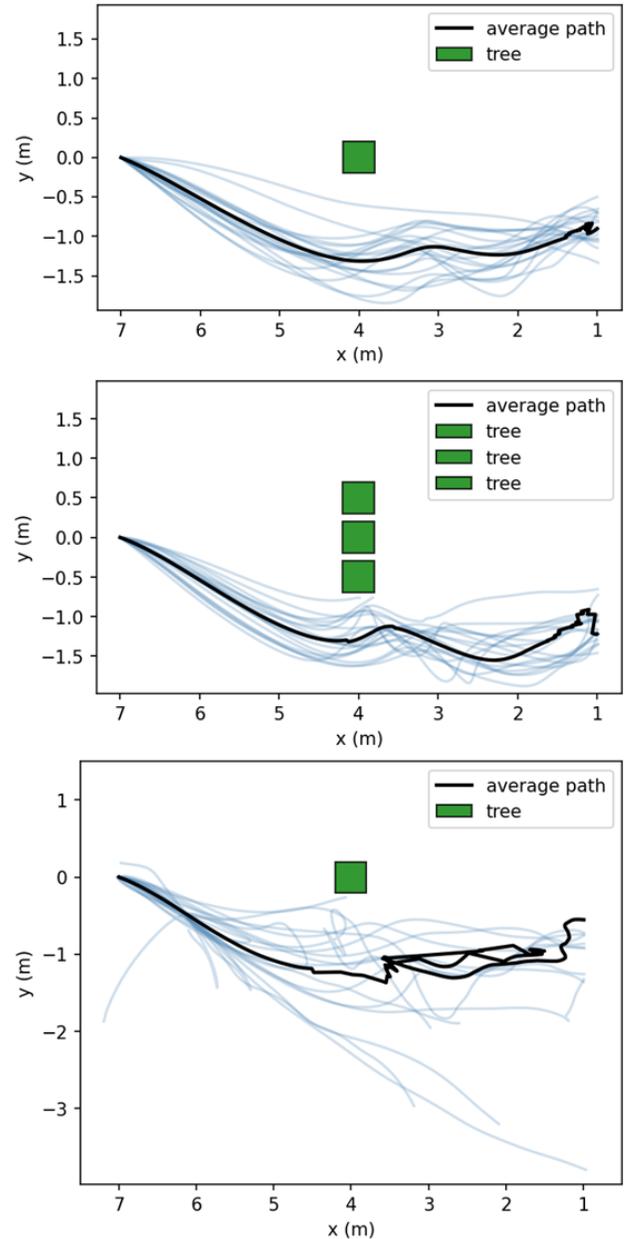

Figure 5: Obstacle avoidance trajectories for 20 rollouts for the (top) single tree cluster with no wind, (middle) three tree cluster and no wind, and (bottom) single tree cluster with medium wind.

in which the control action is applied with both too low and too high magnitude.

### 4.6 Derivation of Safety Requirements

Based on the results from the initial training and subtask evaluation, it can be concluded that the RL agent is safe for deployment under conditions that closely match the baseline training protocol. This includes obstacles that do not vary in size and wind conditions up to 4.5 m/s. Limiting the operational bounds ensures the success of the mission and minimization of loss scenarios. However, if a more general RL agent is desired, additional modifications are required.

The training protocol can be improved by implementing additional countermeasures that improve the RL agent's robustness to environmental perturbations. The subsequent phase discusses the design and implementation of these training improvements. However, once all safety requirements are satisfied and the RL agent is deemed appropriate for deployment, the iterative portion of the RL-STPA process is terminated.

### 4.7 Derivation of Safety Requirements

During the final phase, the identified countermeasures are then incorporated to enhance the training protocol and the update reward function to capture behaviors that led to failure. In this scenario, it was determined that the training curriculum was lacking in exposure to environmental perturbations. Therefore, additional training phases have been identified to introduce wind at increasing levels to improve the ability of the drone to navigate around obstacles when influenced by outside forces. After the initial training, curriculum stages were added that train the RL agent under each wind level for 3 million time steps per stage. The minimum separation requirement was also not strictly adhered to, so the reward function was updated to penalize the RL agent for exceeding the minimum distance to obstacles it encounters.

Due to space constraints, the results of the iterative implementation are not presented but will be explored further in future work.

## 5. Discussion and Limitations

RL-STPA provides several advantages over existing safety analysis methods:
- **Systematic Coverage**: RL-STPA systematically explores the state-action space through perturbation analysis. The subtask decomposition ensures all mission phases are equally explored.
- **Emergent Behavior Detection**: By iteratively testing and improving the learned policy RL-STPA can identify loss scenarios arising from learned feature interactions and insufficient reward functions.
- **Actionable Outputs:** The framework produces concrete safety requirements and training modifications, creating a clear path from analysis to improvement.

Despite its promise, RL-STPA faces several limitations:
- **Manual Subtask Definition:** Current implementation relies heavily on domain expertise for subtask decomposition.
- **Black-Box Interpretation:** While we can identify statistical correlations between conditions and failures, true causal understanding of neural network decisions remains limited.
- **Completeness Guarantees:** Unlike formal verification, RL-STPA cannot guarantee identification of all possible loss scenarios. Perturbation testing, however systematic, may miss rare edge cases.
- **Computational Costs:** Comprehensive perturbation analysis requires substantial computational resources, especially for high-dimensional continuous spaces.
- **Sim-to-Real Gap:** Our analysis assumes simulation fidelity. Real-world deployment may reveal loss scenarios that are not capturable in simulation.

## 6. Conclusion and Future Work

This paper introduced RL-STPA, a framework adapting System-Theoretic Process Analysis for the unique challenges of reinforcement learning in safety-critical applications. By combining subtask decomposition, systematic perturbation testing, and iterative training refinement, RL-STPA provides a practical methodology for identifying and mitigating hazards in learned policies. Our drone navigation case study demonstrates how RL-STPA has the potential to reveal loss scenarios that standard evaluation protocols might miss. The systematic approach uncovered critical vulnerabilities in obstacle avoidance behavior under wind perturbations, insights that directly translate to safety requirements and training improvements. While RL-STPA cannot provide formal guarantees, it offers a structured process for building confidence in safety-critical RL deployments. As RL expands into domains where failures carry severe consequences, frameworks like RL-STPA become essential for responsible deployment.

Several future research directions have been identified that could enhance RL-STPA:
- **Automated Subtask Discovery:** Develop unsupervised learning methods to automatically identify meaningful subtasks from policy behavior.
- **Formal Integration:** Combine RL-STPA with formal verification techniques for stronger guarantees in critical regions of the state space.
- **Active Perturbation Selection:** Replace grid search with adaptive methods that focus computational resources on high-risk regions.
- **Online Adaptation:** Extend RL-STPA for runtime monitoring and dynamic safety bound adjustment.

## Acknowledgments

The work presented here is funded by the Army through STTR-A22B-T002 "Metrics and Methods for Verification, Validation, Assurance and Trust of Machine Learning Models & Data for Safety-Critical Applications in Armaments Systems" (Contract #W15QKN-24-C-0038).